# Multi-Task Deep Learning for Legal Document Translation, Summarization and Multi-Label Classification

Ahmed ELNAGGAR [1], Christoph GEBENDORFER [a], Ingo GLASER [a], and Florian MATTHES [a]

[a] *Software Engineering for Business Information Systems, Technische Universität München, Germany*

**Abstract.** The digitalization of the legal domain has been ongoing for a couple of years. In that process, the application of different machine learning (ML) techniques is crucial. Tasks such as the classification of legal documents or contract clauses as well as the translation of those are highly relevant. On the other side, digitized documents are barely accessible in this field, particularly in Germany. Today, deep learning (DL) is one of the hot topics with many publications and various applications. Sometimes it provides results outperforming the human level. Hence this technique may be feasible for the legal domain as well. However, DL requires thousands of samples to provide decent results. A potential solution to this problem is multi-task DL to enable transfer learning. This approach may be able to overcome the data scarcity problem in the legal domain, specifically for the German language. We applied the state of the art multi-task model on three tasks: translation, summarization, and multi-label classification. The experiments were conducted on legal document corpora utilizing several task combinations as well as various model parameters. The goal was to find the optimal configuration for the tasks at hand within the legal domain. The multi-task DL approach outperformed the state of the art results in all three tasks. This opens a new direction to integrate DL technology more efficiently in the legal domain.

## 1. Introduction

On the past few years, deep learning yielded to great results in many fields, including computer vision, natural language processing (NLP) , speech and robotics. In many areas, it was able to out-perform humans including, image classification [1], health [2] and reading comprehension [3]. The availability of large amount of annotated data and fast computing power are the two main reasons behind this big hype. In the legal domain, legal professionals are doing a lot of tasks related to natural language processing daily, which could be replaced by ML algorithm, but that didn't happen yet deeply because the scarcity of annotated data. Despite that fact that there are exceptionally large text base in the legal domain, it was not preprocessed and structured in a format that could be used for ML technology. The use of ML in the legal domain started to take the attentions of the legal profession and some work has already been done, like translating legal documents [4] or classifying verdicts of the French Supreme Court [5].However, a a lot of possible use cases are not exploited yet.

[1]Corresponding Author: Ahmed Elnaggar, Software Engineering for Business Information Systems, Boltzmannstr. 3, 85748 Garching bei München, Germany; E-mail: ahmed.elnaggar@tum.de.

Generating annotated datasets is generally a costly process. It is even more difficult in the legal domain because we can't easily crowd source it. For example "image net" the biggest image classification dataset and "SQUAD" the biggest reading comprehension dataset , were created through Amazon Mechanical Turk, by sourcing people without very specific knowledge. In the legal domain, we need people with very specific knowledge and education to annotate these unstructured data. Which is hard to crowd source it and even more costly. This leads to very circuital problem on the legal domain:

- NLP is highly required for the legal domain, but the annotated datasets barely exist at all.

One way to overcome this problem is by using multi-task deep learning [6]. In this approach, we train multiple tasks using only one model to provide better results of these problems through transfer learning, especially, tasks that suffers from data scarcity. Therefore, in our work, we needed to achieve two goals:

1. Investigate the effect of transfer learning in the legal problems.
2. Find a big legal text dataset that could be used for transfer learning for any other legal task.

Furthermore, we want to answer three questions regarding the usage of the multi-task deep learning in the legal domain:

1. Is transfer learning through multi-tasking benefits tasks in the legal domain?
2. What are the results of training multiple problems jointly versus separately?
3. Can the multi-task approach outperform the state of the art in the legal domain?

## 2. Related Work

The usage of deep learning has not used intensively in the legal domain. Furthermore, according to our knowledge, the multi-task deep learning was not deeply investigated by researchers and has not been applied in the legal domain. However, we will try to cover the most related research to our work.

**Translation:** A. Vaswani [7] proposed the transformer which represents the current state of the art in general translation, with a BLEU [8] score of 41.8. P.Koehn [4] built 462 machine translation systems for all language pairs of the Acquis Communautaire corpus, which is the body of common rights and obligations which have been adopted by all European Union (EU) Member States.

**Summarization:** AM. Rush [9] initiated work on abstractive summarization with neural networks and induced researchers to continue with sequence-to-sequence models. Additional variants were proposed after that for both extractive and abstractive summarization [10]. C. Grover [11] build the HOLJ corpus for extractive summa-rization of British judgments. B. Hachey [12] used machine learning for extractive summarization using a corpus of judgments of the UK House of Lords.

**Classification:** [13] Multi-label classification of legal document of the JRC-Acquis using the EuroVoc thesaurus [14,15] is one of the difficult tasks because it has more than 6000 labels and low number of samples per label. R. Steinberger [13] achieved a respectable accuracy of 47.3% on German and 48% on English documents of the JRC-Acquis involving the EuroVoc thesaurus.

**Multi-Task:** R. Collobert [16] build a unified multi-task architecture for various NLP tasks such as SRL, NER, POS, chunking and language modeling. They demonstrated that learning tasks simultaneously can improve performance, and they achieved state-of-the-art performance in SRL by training the SRL task jointly with language model. X. Liu [17] successfully develop a multi-task DNN to combine tasks as disparate as classification and web page ranking. The experimental results demonstrate that the model consistently outperforms strong baselines. P. Liu [18] proposed three RNN based architectures to model text sequence with multi-task learning. They focused their work on four different text classification tasks about movies reviews. H. Zhang [19] proposed a multi-task learning architecture for text classification with four types of recurrent neural layers. Their model outperforms the single task models for various datasets for products and movies reviews. L. Kaiser [20] took the next step of multi-task learning by combining tasks from different modalities including image classification, image caption generation, text translation, text parsing and speech recognition. They showed that adding these tasks together never hurts performance and in most cases improves it on all tasks. They also showed that tasks with less data benefit largely from joint training with other tasks, while performance on large tasks degrades only slightly if at all.

## 3. Legal Corpora

The three datasets that were used are the proceedings of the European parliament (Europarl) [21], digital corpus of the European parliament (DCEP) [22] and Joint Research Centre - Acquis Communautaire (JRC-Acquis) [23].

The Europarl corpus provides the proceedings of the European Parliament between the years 1996 and 2011 for 20 languages. Usually, the documents cover the discussions of political topics. Frequently, sentences contain first-person narrative text expressing political opinions and positions. The DCEP covers different areas including press releases, session protocols, reports of the parliamentary committees and written questions for 23 languages. The JRC-Acquis is a collection of legislative documents, retrieved from the European Union (EU) law, stating EU laws and policies for 22 languages, which have to be implemented by each member state.

Only seven major languages were selected for training as proof of concept including English, German, French, Italian, Spanish, Czech and Swedish. Furthermore, the three datasets were preprocessed from their original format to Moses format [24], which ease the integration of any machine learning platform or library.

**Translation Dataset:** The three datasets were used for the translation is Europarl [2], DCEP[3] and JRC-Acquis[4]. Including only the previous mentioned 7 languages. The final combined translation dataset contains 4 to 8 million samples sentences per language pair. It is consider a good source for transfer learning, because Summarization and multi-labeling datasets are only 0.5% and 0.3% of it is size.

**Summarization Dataset:** The JRC-Acquis[5] dataset was used for summarization where each document contains a title element holding a short description of the document body. This summary usually varies between one to three sentences representing the semantic core of each document. The dataset contains between 18k to 22k samples per each language.

**Multi-Labeling Dataset:** The JRC-Acquis[6] dataset was used for Multi-Labeling, where each document is assigned to EuroVoc thesaurus annotations. These EuroVoc thesaurus has a hierarchical structure with over than 6000 classes, for example: agriculture, food, health, information technology, law or politics. Furthermore, Each document is usually assigned to various classes that ranges between one and seven classes. The dataset contains between 11k to 14k samples per each language.

## 4. Multi-Task Legal System

The algorithm we used for multi-task learning is MultiModel algorithm. The algorithm was proposed by Google [20] to create a single generalized deep learning model which is capable of solving tasks across multiple areas (natural language processing, computer vision and speech recognition). This single model was originally trained concurrently on general tasks including image classification, image captioning generation, language translation, English parsing task and speech recognition. However, in our work we used the algorithm for language translation, summarization and document classification specifically in the legal domain.

### *4.1. MultiModel Architecture*

The multi-model uses the sequence to sequence approach based on convolutional neural network. The model consists of four building blocks (Modality Nets, Encoder, Decoder and mixer) as shown on figure 1 and briefly presented [20] bellow.

**Modality Nets:** The model uses four different modality nets (language, image, audio, categorical).This allows it to accept and produce different inputs and outputs types. However, it uses only the language, image and audio for inputs and language and categorical for outputs. Furthermore, it produces a unified representation for all of the tasks. In our case, only the language modality was used.

**Encoder:** The encoder takes the unified output of the modality nets and process it with six custom built convolutional blocks with one mixture-of-expert layer in between.

[2]https://mediatum.ub.tum.de/1446650
[3]https://mediatum.ub.tum.de/1446648
[4]https://mediatum.ub.tum.de/1446655
[5]https://mediatum.ub.tum.de/1446654
[6]https://mediatum.ub.tum.de/1446653

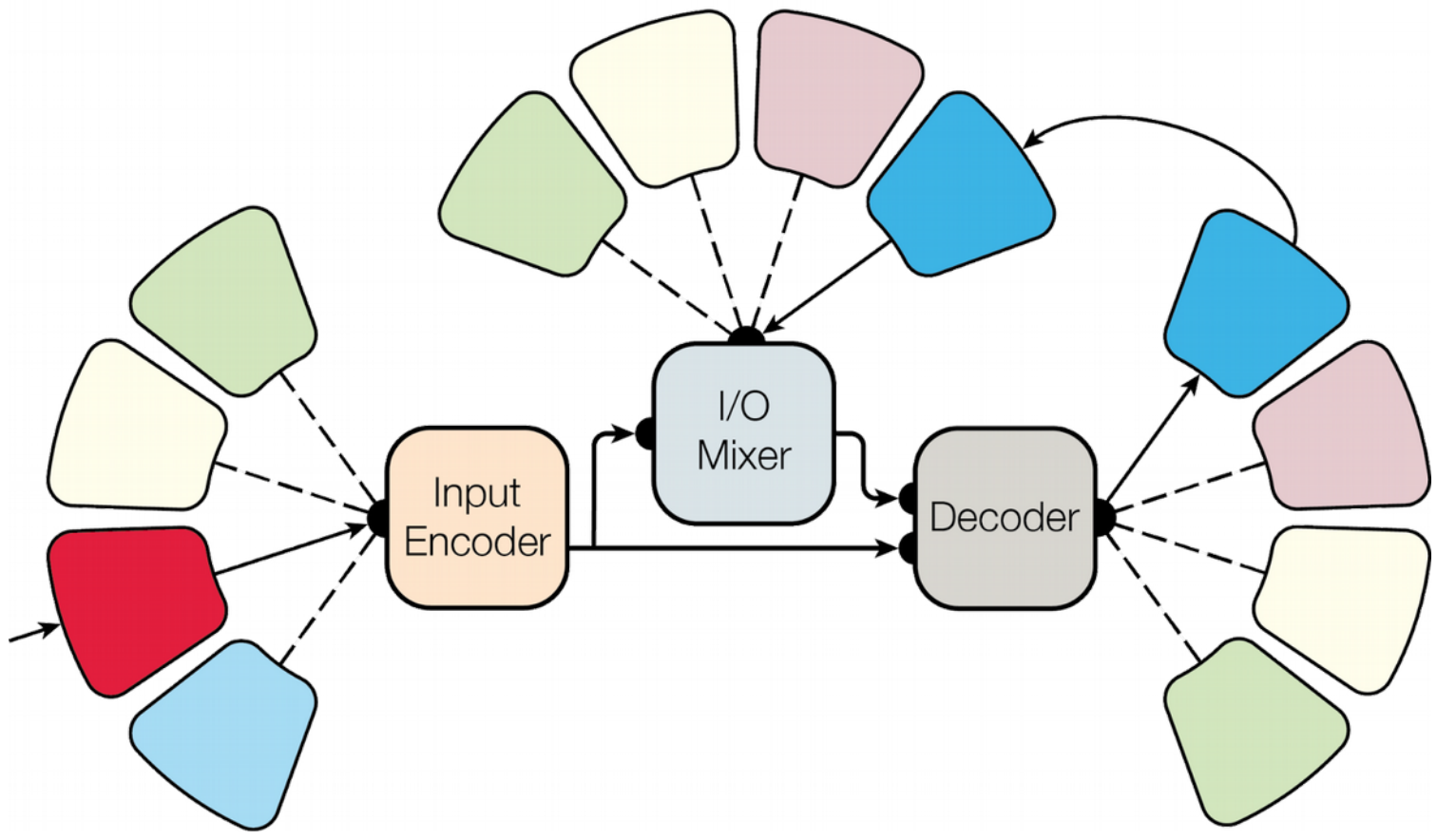

**Figure 1.** Google Multi-Model Building Blocks

**Decoder:** The decoder takes the output from the encoder and produce the final output using modality nets. It consists of four convolutional attention blocks with one mixture of expert layer in the middle. Furthermore, at each training step a token is passed to it, which allows it learn different representations for different tasks.

**I/O Mixer:** The mixer takes the output of the encoder and the previous output from the decoder. This allows it to learn long term dependencies. It consists of two convolutional blocks and one attention block.

## 5. Experimental Settings

### *5.1. Training Details*

Generally, every model was trained until it converged, and sometimes we used early stop to prevent over-fitting. For the multimodel, we have used two configurations the base (MM-B) configuration as was described in the paper and light version (MM-L) configuration. The light version has fewer parameters and was used to test the effect of number of parameters on the result. The transformer, multi-model base and multi-model light were trained with batch size 2048, 2048 and 1048, while the hidden size of each layer was 128, 512 and 512, and the filter size was 1024, 2084 and 2048. In case of our Multi-Task model at each training step we trained the model for the same batch size of each problem sequentially. To speed up the process, we trained the algorithms on four machines. The transformer model on a machine with 4x Tesla K80, the multimodel base version on two machines the first was NVIDIA DGX-1 with 8x Tesla V100 and the second with 5x Titan XP, and the multimodel light version on a machine with 4x Titan 1080Ti.

Different combinations of the jointly tasks has been tested. For translation, we choose two combinations, the pool combination (jt-pool-5) consists of the five available German translation pairs ”de-en, de-es, de-fr, de-it, de-sv”, and the chain combination (jt-chain-7) which consists of a chain of language ”cs-de, de-en, en-es, es-fr, fr-it, it-sv”. For summarization, we had one combination (js-7) which joint all the summarization languages. For multi-labeling, we had one combination (jl-7) which joint all the multi-labeling languages. Finally, we had a last combination which combines different tasks with the same language (ja-3). It combines the translation, summarization and multi-labeling tasks of the same source language together. All of these combinations of the Multi-Model was compared with the result of the state of the art models, which is the transformer for general translation and summarization, and JEX for JRC-Acquis multi-label classification. Finally, due to the time and the number of pages constrains we only report the result of the German language.

## *5.2. Metrics*

We report our results with common task-dependent metrics. In the follow sections we cover each task metrics.

### *5.2.1. Translation*

The BLEU [8] score was used to evaluate the translation results. It measure the quality of the translation based in the n-grams overlaps between the predicted translation and the target translation.

$$BLEU = min(1, \frac{hypothesis_length}{reference_length})(\prod_{i=1}^{4} precision_i)^{\frac{1}{4}} \quad (1)$$

### *5.2.2. Summarization*

The standard metric for evaluating the summarization is ROUGE [25] score, which we used in for the summary evaluation. We only evaluated the results based on 1-gram, 2-grams and the longest n-gram. They simply called ROUGE-1, ROUGE-2 and ROUGE-L.

$$ROUGE_N = \frac{\sum_{S \in reference_summaries} \sum_{gram_n \in S} count_match(gram_n)}{\sum_{S \in reference_summaries} \sum_{gram_n \in S} count(gram_n)} \quad (2)$$

### *5.2.3. Multi-Label Classification*

For multi-label classification, we report precision, recall and F1 score.

$$Precision = \frac{\#TruePositive}{\#TruePositive + \#FalsePositive} \quad (3)$$

$$Recall = \frac{\#TruePositive}{\#TruePositive + \#FalseNegative} \quad (4)$$

$$F_1 = \frac{Precision \times Recall}{Precision + Recall} \quad (5)$$

## 6. Result and Discussion

Figure 2 shows the translation results. Generally, all models had better result on both dcep and jrc-acquis datasets than the europarl. This might be because these two datasets contain a lot of cross references, sentence fragments and enumerations compared to the europarl.

The Multi-model light version (MM-L single, MM-L jt-pool-5, MM-L jt-chain-7 and MML- ja-3) falls behind both the transformer model (TF-B single) and the multi-model base version on the three datasets. This because the number of parameters of the light version is almost half of the number of parameters of the other models. The light version usually produces shorter sentences, however, after manually examine them, we found that semantic meaning remains largely untouched. Another observation that multi-model light single which was trained on a single task outperform the same model but with joint tasks. This is because the limited capacity of the model didn't allow it to learn multi-tasks jointly. By increasing the number of tasks the BLEU score decrease.

The Multi-model base version (MM-B single) outperformed the transformer model for both dcep and jrc-acquis datasets with BLEU score 54.98 and 67.24 compared to 53.3 and 64.22. However, in the case of europarl dataset, the BLEU score was a little bit less, 37.15 compared to 37.34. When the model was trained jointly with other translation languages (MM-B jt-pool-5 and MM-B ja-chain-7) the BLEU score falls behind the transformer. However, the model (MM-B ja-3) which was trained with different tasks (summarization and classification) with the same input language (Germany), outperformed all other models in the dcep dataset with BLEU score 55.11. It was better than the transformer, but less than the multi-model which was trained on a single task (MM-B single) for the acquis with BLEU score 66.6. For the europarl dataset it slightly falls behind both the transformer and the multi-model base single task.

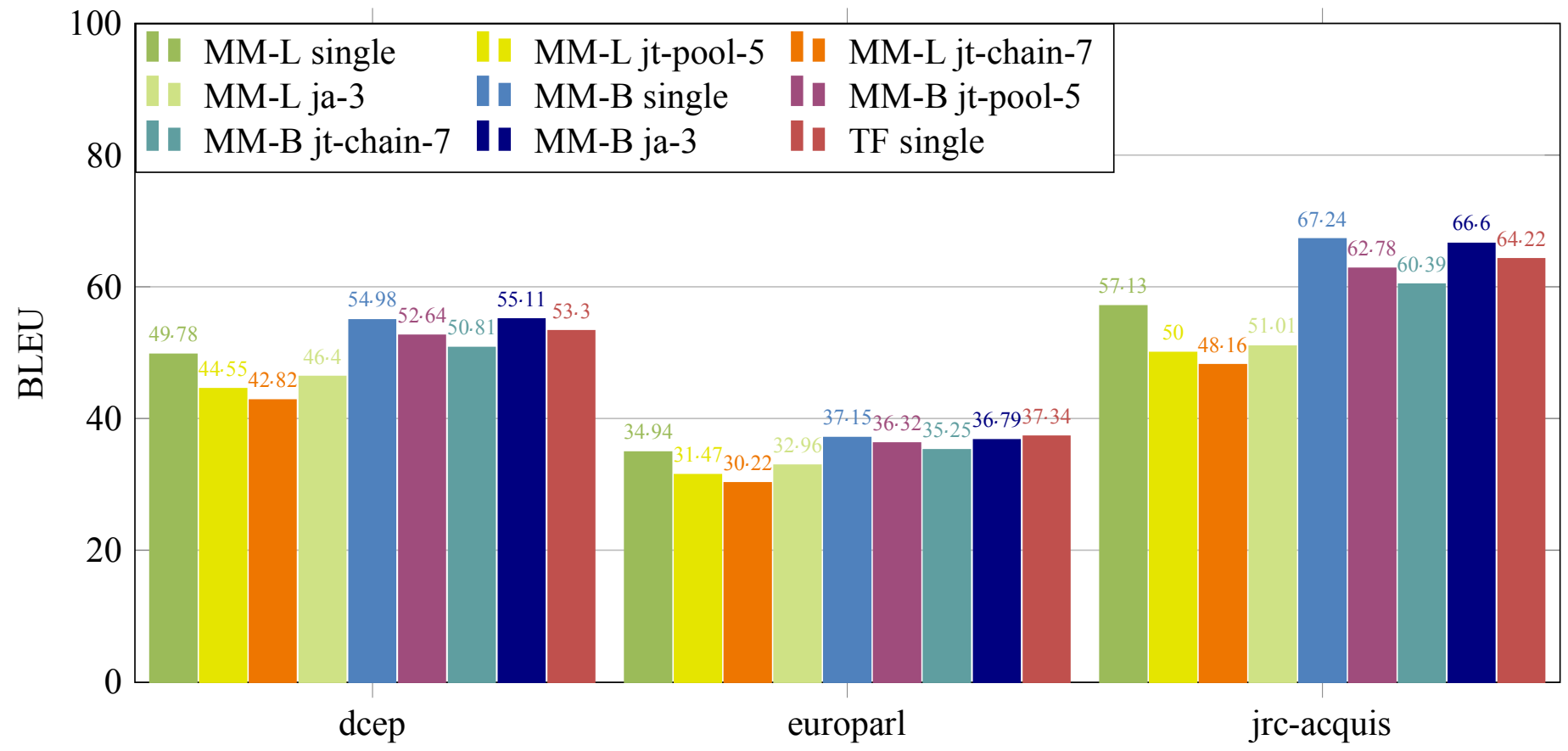

**Figure 2.** German-to-English translation BLEU score performance for all single-task & multi-task translation combinations trained on the MultiModel Light (MM-L), MultiModel Base (MM-B) and Transformer (TF)

Figure 3 shows the summarization results. The transformer model falls behind the multi-model for both the light and base versions. It had almost 50% of the ROUGE of the multi-model. The multi-model light versions had almost always better ROUGE scores than the multi-model base when it trained on either single German summarization or multi-language summarization. The reason is that the dataset of summarization was small relative to the number of parameters for the base model, which lead to fast over-fitting even with using regularization techniques. The best ROUGE scores were obtained from the multi model (MM-B ja-3) which was trained on the three different tasks jointly with the same input language (Germany), with ROUGE-1, ROUGE-2 and ROUGE-L of 0.82, 0.75 and 0.82.

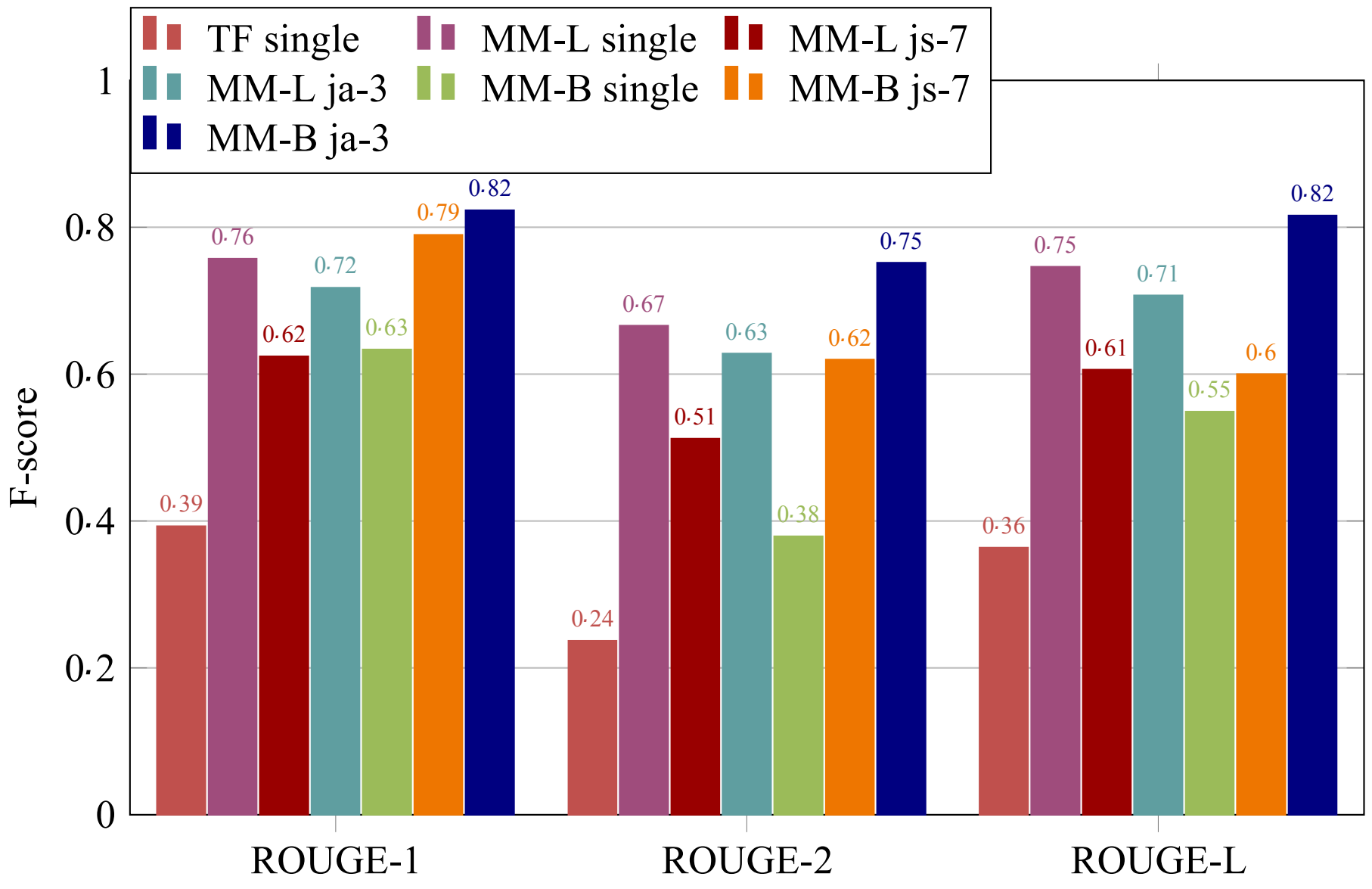

**Figure 3.** German Summarization performance using ROUGE score for all single-task & multi-task translation combinations trained on the MultiModel Light (MM-L), MultiModel Base (MM-B) and Transformer (TF)

Figure 4 shows the multi-label classification results. The JEX model out-performed the multi-model light (MM-L single, MM-L jl-7 and ML-L ja-3)on both F-score and recall, but it had lower precision . The multi-model base which was trained on single (MM-B single) and all classification languages (MM-B jl-7) falls badly to provide any good classification. The reason behind that is the multi-labeling datasets, which is very small compared to the model capacity, that made the model to over-fit. The best result that out-performed JEX the state of the art model was obtained by combining the three different tasks together with the same language (MM-B ja-3) with F-score, recall and precision of 0.65, 0.63 and 0.67 compared to 0.51, 0.55 and 0.47.

The previous experiments lead to three important points, which answers the three research questions. First, multi-task deep learning out-perform the single task state of the art models, when it is combined with different tasks of the same input language and one of these tasks has a large number of samples. This allows to transfer the knowledge the algorithm learn between these tasks. Second, the greater the number of tasks in a joint task the greater the impact on performance than the relatedness or diversity of the joined tasks. Third, the capacity of the multi-task models must be adopted depending on datasets sizes [7].

[7] The output of the translation, summarization and classification tasks with the different models can be downloaded from [2], [3] and [4] in the decodes folder.

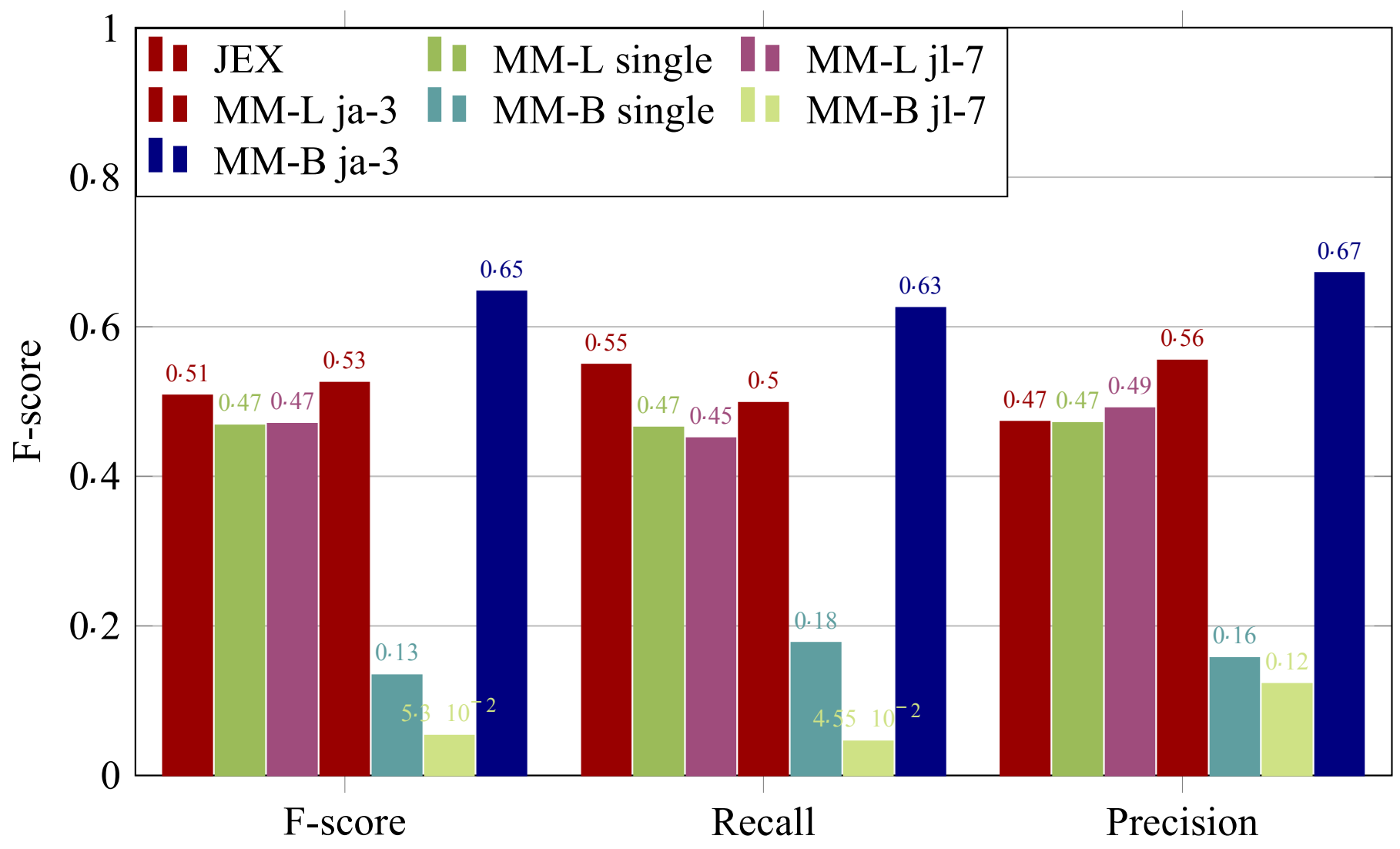

**Figure 4.** German multi-label classification performance using F-score, Recall and Precision scores for all single-task & multi-task translation combinations trained on the MultiModel Light (MM-L), MultiModel Base (MM-B) and JEX [26]

## 7. Conclusions & Future Work

We proved that multi-task deep learning can be useful in the legal domain. Of course, the type, the amount of joined tasks and the capacity of the multi-task model are major Influential factors of the result. However, It is an effective approach to solve the data scarcity problem through transfer learning. Using this approach will allow us to outperform the current state of the art results, and allow the usage of the deep learning technology on the legal domain. Our work is a base for further research on the effectiveness and usage of multi-task in the legal domain. However, more experiments are required to test it on other tasks, datasets, languages and training combinations. The provided datasets could be used to test the approach on the rest six languages.

## 8. Acknowledgements

We gratefully acknowledge the support of Leibniz-Rechenzentrum, Microsoft Corporation and NVIDIA Corporation with hardware which were used for this research.